\documentclass[letterpaper, 10 pt, conference]{ieeeconf}  

\IEEEoverridecommandlockouts                              

\usepackage{graphics} 

\usepackage{hyperref}
\hypersetup{
    colorlinks=true,
    linkcolor=black,
    filecolor=magenta,      
    urlcolor=blue,
    pdftitle={Overleaf Example},
    pdfpagemode=FullScreen,
    }

\usepackage{graphicx}
\usepackage[capitalise,nameinlink]{cleveref}
\usepackage{subfig}

\usepackage{stfloats}

\usepackage[table,xcdraw]{xcolor}

\title{\LARGE \bf
UrbanNet: Leveraging Urban Maps for Long Range 3D Object Detection
}

\author{Juan Carrillo and Steven Waslander \\
  University of Toronto

\thanks{Juan Carrillo and Steven Waslander are with the Institute for Aerospace Studies, University of Toronto, 4925 Dufferin
St, North York, ON, Canada.
        {\tt\small juan.carrillo@mail.utoronto.ca, }
        {\tt\small stevenw@utias.utoronto.ca}}%
}

\begin{document}

\maketitle
\thispagestyle{empty}
\pagestyle{empty}

\begin{abstract}

Relying on monocular image data for precise 3D object detection remains an open problem, whose solution has broad implications for cost-sensitive applications such as traffic monitoring.  We present UrbanNet, a modular architecture for long range monocular 3D object detection with static cameras. Our proposed system combines commonly available urban maps along with a mature 2D object detector and an efficient 3D object descriptor to accomplish accurate detection at long range even when objects are rotated along any of their three axes. We evaluate UrbanNet on a novel challenging synthetic dataset and highlight the advantages of its design for traffic detection in roads with changing slope, where the flat ground approximation does not hold. Data and code are available at \href{https://github.com/TRAILab/UrbanNet}{https://github.com/TRAILab/UrbanNet}
\end{abstract}

\section{INTRODUCTION}
\label{sec:intro}

Vehicular traffic continues to increase each year as cities grow around the globe, leading to challenges related to road safety and efficient mobility. Traffic accidents are among the top causes of death in many countries and road intersections are places of particularly high risk for such unwanted events. According to road safety statistics, intersections are the locations where about four of every 10 vehicle crashes take place in the United States \cite{u2010crash}, and is also where one third of fatalities and 40\% of major lesions happen in Canada \cite{roadcan}. As a result, transportation offices at city governments have been looking at cutting-edge technologies to help prevent accidents at intersections while also improving traffic flow. Dynamic control of traffic lights based on real-time assessment of vehicles at intersections has shown to significantly reduce delays, with examples such as the City of Los Angeles and its Automated Traffic Surveillance and Control (ATSAC) System \cite{lasig2016}, which lowers travel time by as much as 32\%.

Smart intersections are an initiative gaining momentum globally which entails the use of advanced technologies to improve road safety as well as traffic flow \cite{amirgholy2020optimal}. One of the fundamental components in most of these systems is a mechanism to dynamically detect the location of vehicles, cyclists, and pedestrians, among other road users based on raw data collected from a variety of sensors including cameras, LiDAR, RADAR, and inductive-loop detectors. Object detection is the first step to enable a sequence of downstream tasks such as counting and tracking, as well as dynamic control of signals to lower the risk of accidents and optimize vehicle flow. In other words, smart intersections offer challenging scenarios to incorporate recent techniques in Computer Vision and Machine Learning that benefit from large amounts of data and are particularly well suited for complex tasks such as object detection.

\begin{figure}[!htb]
  \includegraphics[width=.47\textwidth]{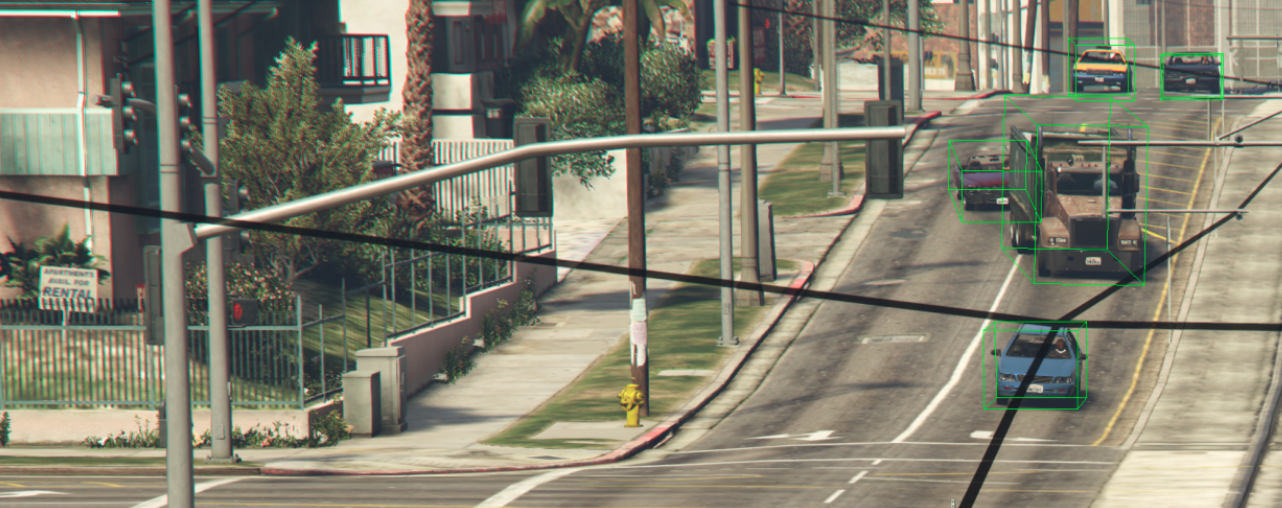}
  \caption{Representing vehicles as 3D bounding boxes involves the definition of their size, location, and orientation.}
  \label{fig:3d_detection_its}
\end{figure}

Object detection tasks are broadly categorized into 2D or 3D detection, with the second group being more challenging due to the increased complexity of the inputs and expected outputs. While competitive 2D object detection methods commonly require only images as inputs \cite{wu2019detectron2}, state of the art 3D object detectors require additional input features to properly locate objects in 3D space.  In particular, range information is often drawn from raw LiDAR point clouds, radar returns or dense depth maps from stereo vision \cite{pang2020clocs}. Moreover, methods that only rely on monocular images for 3D object detection usually show a significant drop in performance as opposed to those using direct measurement of distance to object in the scene. An object is often represented in three dimensions as an oriented 3D bounding box, where at least nine parameters are needed to fully characterize the size and pose of such box. Three parameters for size (height, width, length), three for position (x,y,z), and three for orientation (pitch, yaw, roll). 

Depending on the specific application, we can approximate the road as a flat ground plane across the scene and therefore assume the pitch and roll  rotations relative to the ground plane to be close to zero, making the problem significantly less complex \cite{Geiger2012CVPR}. In other words, assuming a flat ground plane provides a good approximation for close range objects that is also convenient when the road plane is parallel with the plane formed by two axis in the coordinate reference system of one or more sensors. In contrast, for long range 3D object detection using traffic cameras at intersections, the flat ground plane assumption is regularly invalidated, and it becomes essential to account for varying slope to correctly characterize the size and pose of each object. \Cref{fig:3d_detection_its} shows an example of a 3D bounding box projected on a traffic camera image along with its representation in a 3D coordinate frame. As a result, this work emphasizes the detection of a full 3DOF orientation estimate in the form of three Euler angle parameters for each detected bounding box.

State of the art 3D object detection methods, studied primarily in the context of autonomous driving, make extensive use of LiDAR data to obtain accurate spatial features. Similarly, 3D maps are rich sources of spatial data useful to complement streams coming from sensors. These accurate maps are created during the development of urban infrastructure projects and are therefore commonly available for most cities. The level of detail in these maps vary from simple elevation maps with contour lines to highly detailed and realistic 3D representations of most elements in urban areas \cite{soon2017citygml}. As a result, more attention has been given lately to urban maps as additional data sources and datasets published recently for research purposes in autonomous driving now incorporate some form of map \cite{Xie2016CVPR}\cite{nuscenes2019}\cite{tan2020toronto}.

In this paper we aim to bridge the gap between 2D and 3D object detection for applications in long range traffic surveillance by introducing the UrbanNet architecture (\cref{fig:UrbanNet}). Our contributions are summarized as follows:
\begin{itemize}
  \item We design an efficient 3D object descriptor that uses cropped image regions with driving lane maps as useful priors to output vehicle 3D bounding boxes along with size and observation angle values.
  \item We introduce a novel normalization step that facilitates the encoding of vehicle keypoints with respect to its tight 2D bounding box and allows the modular use of any out of the box 2D object detector in the first stage of the architecture.
  \item We incorporate detailed geometric information in the form of a 3D environment map that provides elevation and slope throughout the field of view of the camera. This allows us to accurately and efficiently determine the pose of vehicles at long range.
\end{itemize}

The rest of the paper is organized in the following sections. In \cref{sec:relatedwork} we review previous work in the context of 2D and 3D vehicle detection. In \cref{sec:UrbanNet} we describe in more detail the UrbanNet architecture and in \cref{sec:experiments} we present experiments and results.

\section{Related Work}
\label{sec:relatedwork}

This review of related work focuses mainly on publications targeted at detection of vehicles from sensors installed at intersections; however, methods designed for data coming from sensors onboard autonomous vehicles are also considered.

\subsection{2D Object Detection}

Methods for 2D object detection at intersections are mostly adaptations of methods designed for detection of objects in generic datasets \cite{lin2014microsoft}. However, in recent years large datasets have been published to train and evaluate object detectors specifically for objects at intersections, such as the UA-DETRAC \cite{CVIU_UA-DETRAC}, CityFlow \cite{Tang_2019_CVPR}, and MIO-TCD \cite{luo2018mio} datasets. Examples of such customized methods use Deep Learning architectures like Single Shot Detectors \cite{1cb273417866405db145954c097e05e6} and YOLO detectors \cite{10.1145/3373647}. The development on this front also makes use of conventional Computer Vision techniques such as handcrafted features and optical flow \cite{xun2018congestion}, as well as Gaussian mixture models \cite{premachandra2020detection}. In general, these methods are designed for objects at close range, where a flat layout of the intersection is sufficient to locate moving objects in bird's eye view.

\subsection{Image-based 3D Object Detection}

Previous work focused on 3D object detection at intersections using only images is represented mostly by studies that localize and track vehicles in bird's eye view \cite{Hua_2018_CVPR_Workshops} \cite{liu2020robust}, rather than fully characterizing the size and pose of the objects. Such methods are useful for traffic studies that require counts of vehicle turns and traffic queues, and not necessarily full 3D bounding boxes to represent moving objects. While the sensors' mounting is not the same, methods for 3D object detection in the context of autonomous driving have developed interesting ideas such as carefully designed loss functions \cite{simonelli2019disentangling}, the use of auxiliary CAD models \cite{chabot2017deep}, geometric constraints between 2D and 3D bounding boxes \cite{mousavian20173d}, stereo imaging \cite{pon2020object}, and keypoint detection \cite{liu2020smoke}\cite{li2020rtm3d}. However, most of these methods are designed for 3D detection up to 40m \cite{pon2020object}\cite{CaDDN} and their direct implementation at a longer range produces significantly less accurate results, especially when looking at roads with changing slope.

\subsection{LiDAR and Image 3D Object Detection}

Methods using LiDAR sensors installed at intersections offer more accurate distance measurements and commonly provide 3D detections with a full characterization of the size and pose of objects. Some of the common techniques in these methods include voxel maps along with handcrafted 3D features \cite{7535458} or the combination of background filtering and point cloud clustering \cite{doi:10.1177/0361198119844457}\cite{ZHAO201968}\cite{Wu_2020}. In the context of autonomous vehicles, 3D object detection methods use Deep Learning techniques to achieve significantly higher performance when compared to image only methods \cite{ku2018joint}\cite{pang2020clocs}\cite{vora2020pointpainting}. However, regardless of the sensor mounting (poles or vehicles), the operational range of these methods is often limited to close and mid range distances because of the hardware specifications of the sensors and the sparsity of the resulting point cloud. Besides, LiDAR sensors are more expensive than cameras and this reduces the chances of widespread adoption for many city intersections.

\subsection{Maps in 3D Object Detection}

Methods in this category use one or more map representations to narrow down the vehicle detection task and improve the accuracy of the output localization. Some of the common techniques at intersections take advantage of projective geometry to determine correspondences between image coordinates and bird's eye view coordinates \cite{shi2018geometry}\cite{malinovskiy2009video} as well as extraction of road surface area \cite{song2019vision}. However, these methods do not account for elevation and slope variations in the road, which are even more prevalent at long range. For applications in autonomous driving maps are widely used to inform drivable surface \cite{pmlr-v87-yang18b}\cite{cai2018robust}, road boundaries \cite{8456509}, lane center lines \cite{chang2019argoverse}, and volumetric occupancy grids \cite{ravi2018real}. Most of these methods also rely on the approximation of a flat ground plane in the proximity of the ego vehicle, and therefore only estimate heading instead of the full 3D orientation of the vehicle.

\begin{figure*}[!htb]
  \centering
  \includegraphics[width=1\textwidth]{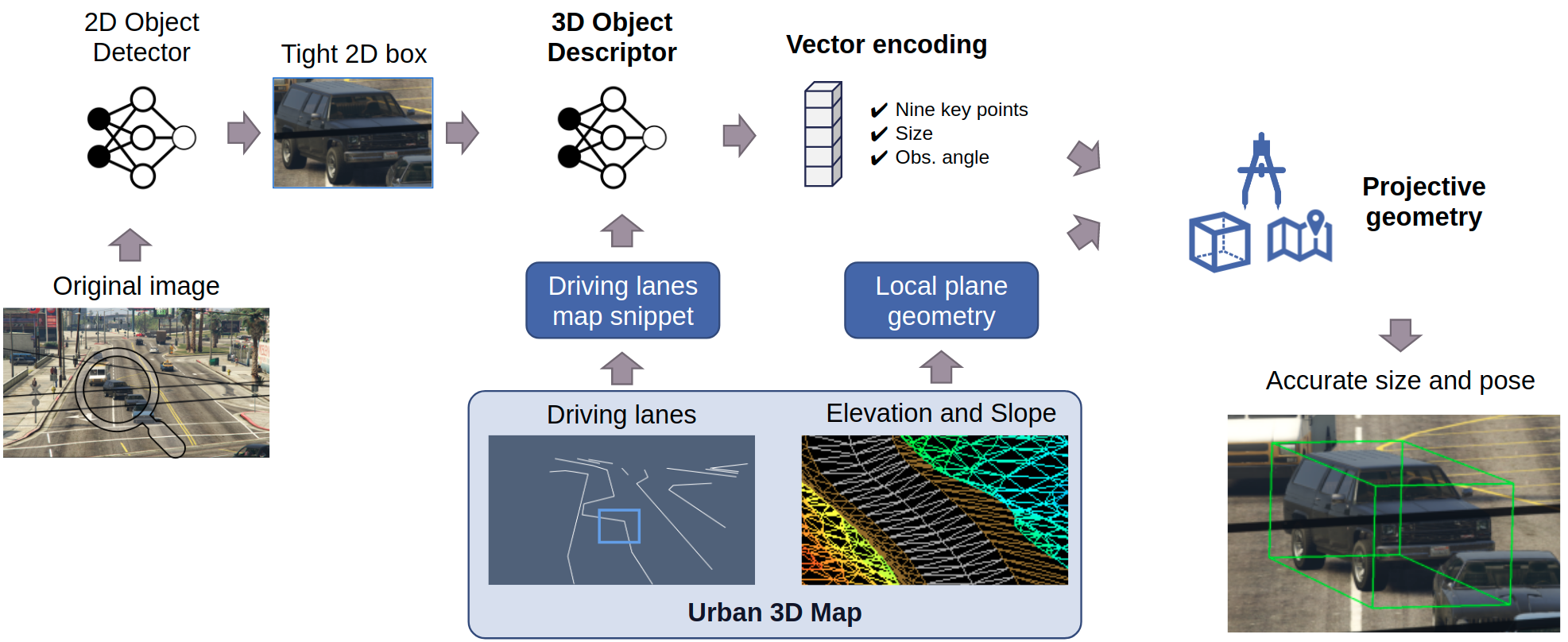}
  \caption{UrbanNet's system overview. The first stage uses a mature 2D object detector to find relevant objects and output a tight 2D bounding box for each. The second stage uses an efficient 3D object descriptor in combination with an urban 3D map to accurately characterize the size and pose of objects in the road.}
  \label{fig:UrbanNet}
\end{figure*}

\section{The UrbanNet Architecture}
\label{sec:UrbanNet}

UrbanNet is a two stage modular architecture designed to achieve an effective balance between accuracy and efficiency for 3D object detection at long range using single camera images along with 3D urban maps (\cref{fig:UrbanNet}). 

\subsection{2D Object Detection}

The first stage uses a mature 2D object detector to process high resolution images and output tight 2D bounding boxes enclosing objects within the classes of interest. While other multi stage architectures require customized 2D bounding boxes that fully enclose the projection of the corresponding 3D box \cite{mousavian20173d}, UrbanNet only needs tight 2D boxes, which allows maximum interoperability with a broad number of pretrained 2D object detectors such as Detectron2 \cite{wu2019detectron2}. Interoperability is key for us because it enables the interchangeable use of highly optimized 2D detectors tailored for edge computing applications. Since the main contributions of our paper are focused on 3D object detection we choose to use ground truth 2D bounding boxes and image snippets for the next stage.

\subsection{Driving lanes map}

\begin{figure*}[!htb]
  \centering
  \subfloat[Sample input image.]{\includegraphics[width=.4\textwidth]{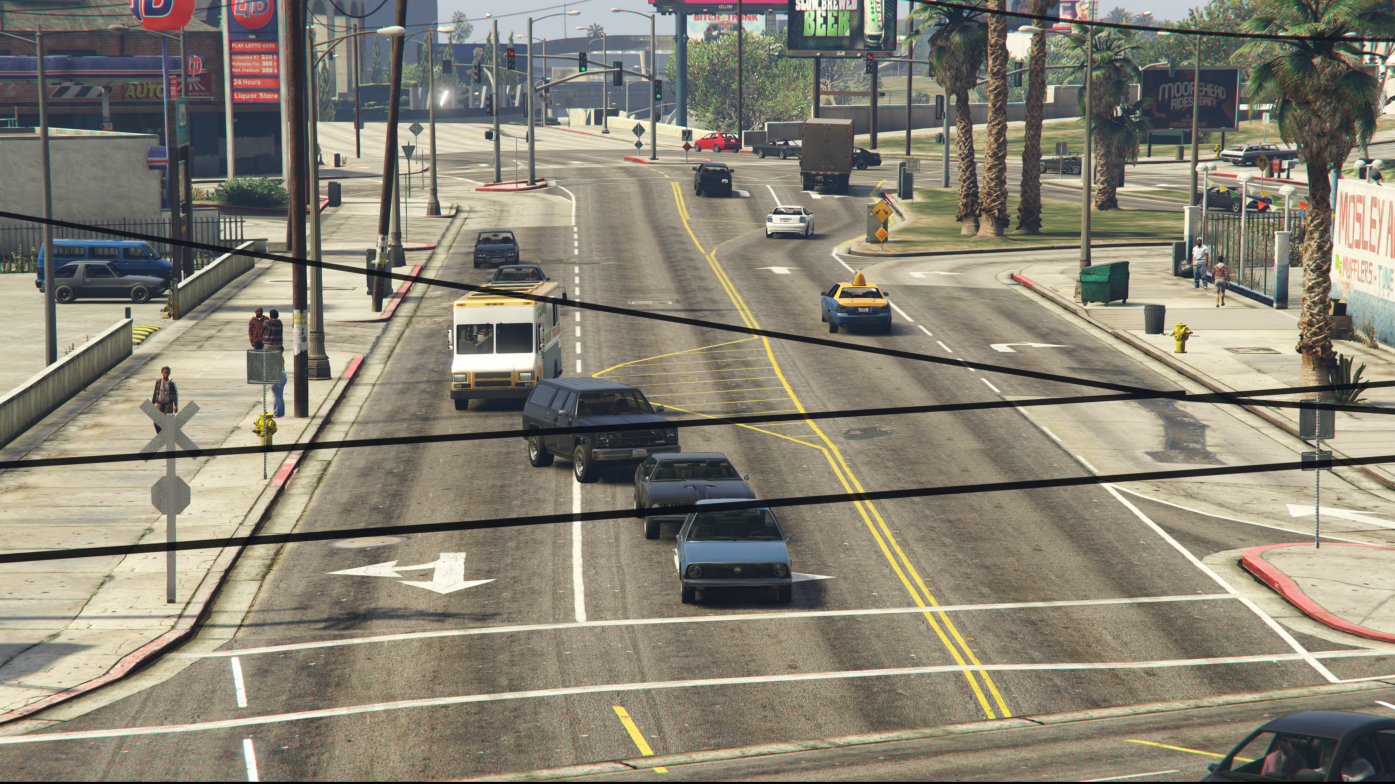}%
    \label{}}
  \hfil
  \subfloat[3D vehicle centroids projected on the road surface along with the lane center lines.]{\includegraphics[width=.4\textwidth]{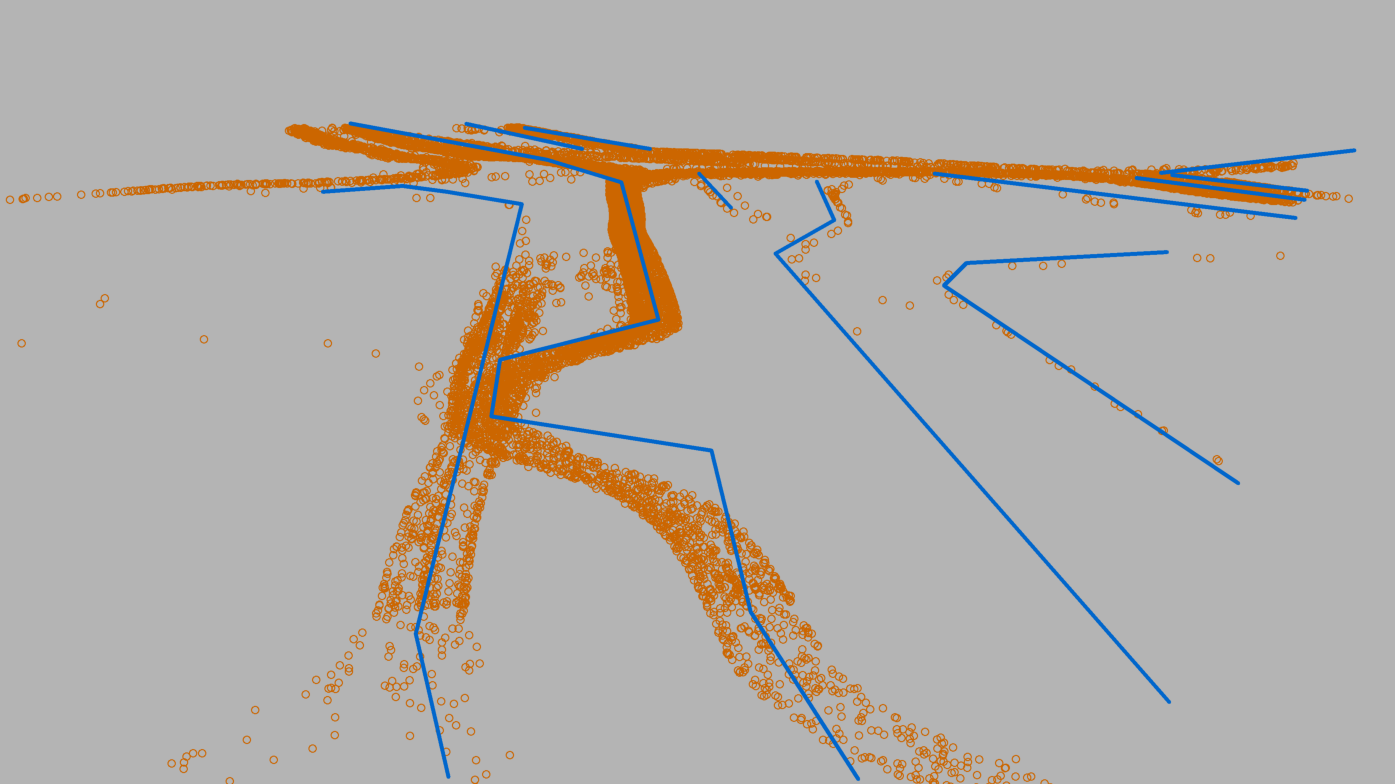}%
    \label{}}
  \caption{A driving lanes map representation at a sample intersection.}
  \label{fig:driv_lanes}
\end{figure*}

Driving lanes are where vehicles are located most of the time and are therefore a useful prior to guide computer vision tasks. More specifically, we explore the use of lane center lines \cite{chang2019argoverse} as an additional feature for the downstream 3D object descriptor. We make a preliminary evaluation of this hypothesis by plotting the projection of 3D vehicle centroids on the road surface along with the lane center lines (\cref{fig:driv_lanes}). Based on this figure we visually confirm the strong spatial correlation between both data sources in this synthetic dataset, and surmize that a similar relationship would hold in real-world data as well. UrbanNet uses center lines as an additional input, which are projected on the image and passed as a fourth channel to the 3D object descriptor. In \cref{sec:experiments} we present ablation experiments to further evaluate the contribution of the centerline map.

\subsection{3D Object Descriptor}

\textit{Model input}

We concatenate the cropped 2D image snippet that tightly encloses the vehicle and the driving lanes into a four channel tensor and pass it as input to the 3D object descriptor. Since image snippets have different sizes depending on the actual dimensions of the objects and how close they are to the camera, all snippets are rescaled and padded with zeros beside their longer edges to fit in a square snippet with predefined height and width.

\begin{figure}
  \includegraphics[width=.48\textwidth]{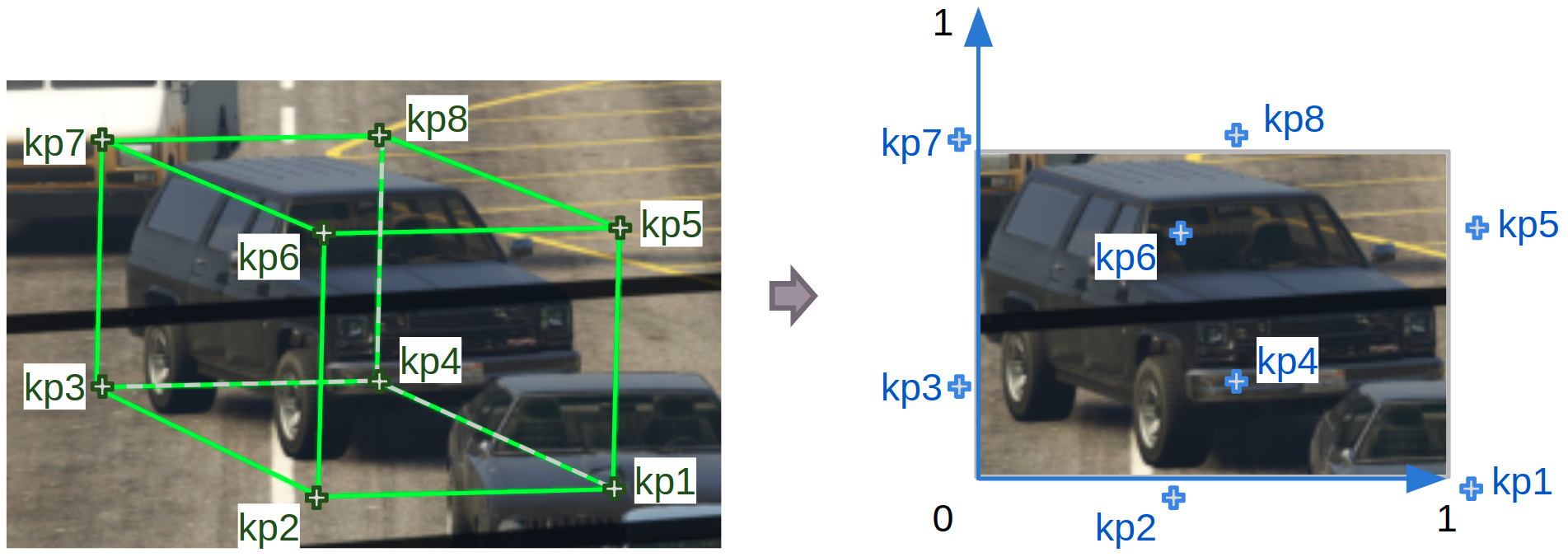}
  \caption{Keypoint normalization.}
  \label{fig:keypoint}
\end{figure}

\textit{Network design}

The  goal of the 3D object descriptor is to accurately determine the keypoints that represent the projection of the object's 3D bounding box into the image 2D space, along with its size and observation angle. We implement this 3D descriptor as an efficient Deep Neural Network inspired by the VGG architecture \cite{simonyan2014very}. Even though more complex and competitive network architectures have been introduced lately for computer vision tasks \cite{10.1007/978-3-030-58520-4_12}, we use a generic network model as the contributions of this work lie in the assessment of benefit of the additional information from urban maps. Our 3D descriptor is made of six consecutive pairs of convolutional and max pooling layers followed by three fully connected layers. Each convolutional and max pooling pair in the backbone reduces by half the height and width of the input tensor and doubles the number of channels. All the convolutional layers have a kernel size of three by three and a stride of one, while all the max pooling layers have a kernel size of two by two and a stride of two. The first and second fully connected layers reduce by four the number of elements in the input tensors, while the third delivers the final output of the network.

\begin{figure*}[!htb]
  \centering
  \subfloat[]{\includegraphics[width=.45\textwidth]{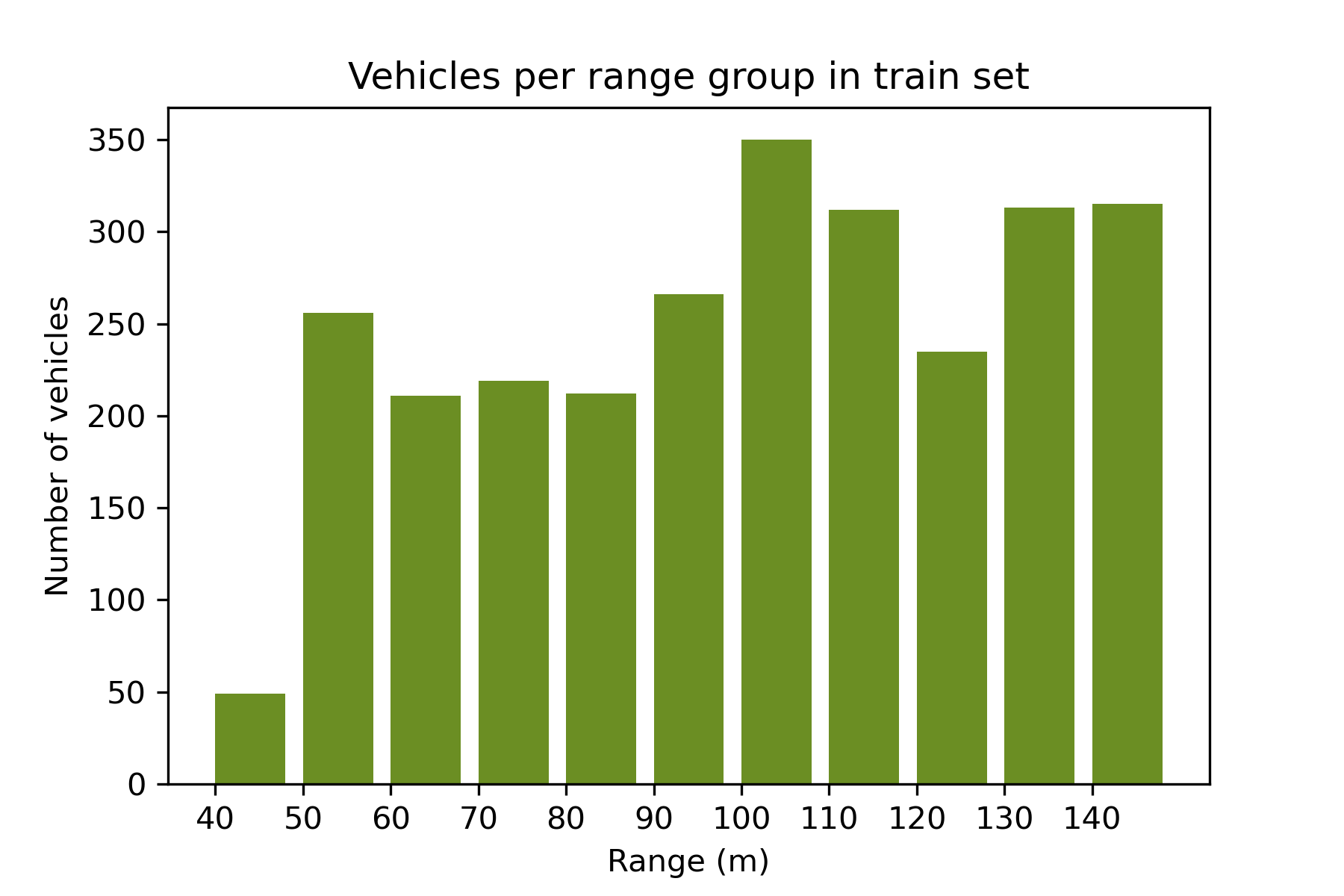}%
    \label{fig_first_case}}
  \hfil
  \subfloat[]{\includegraphics[width=.45\textwidth]{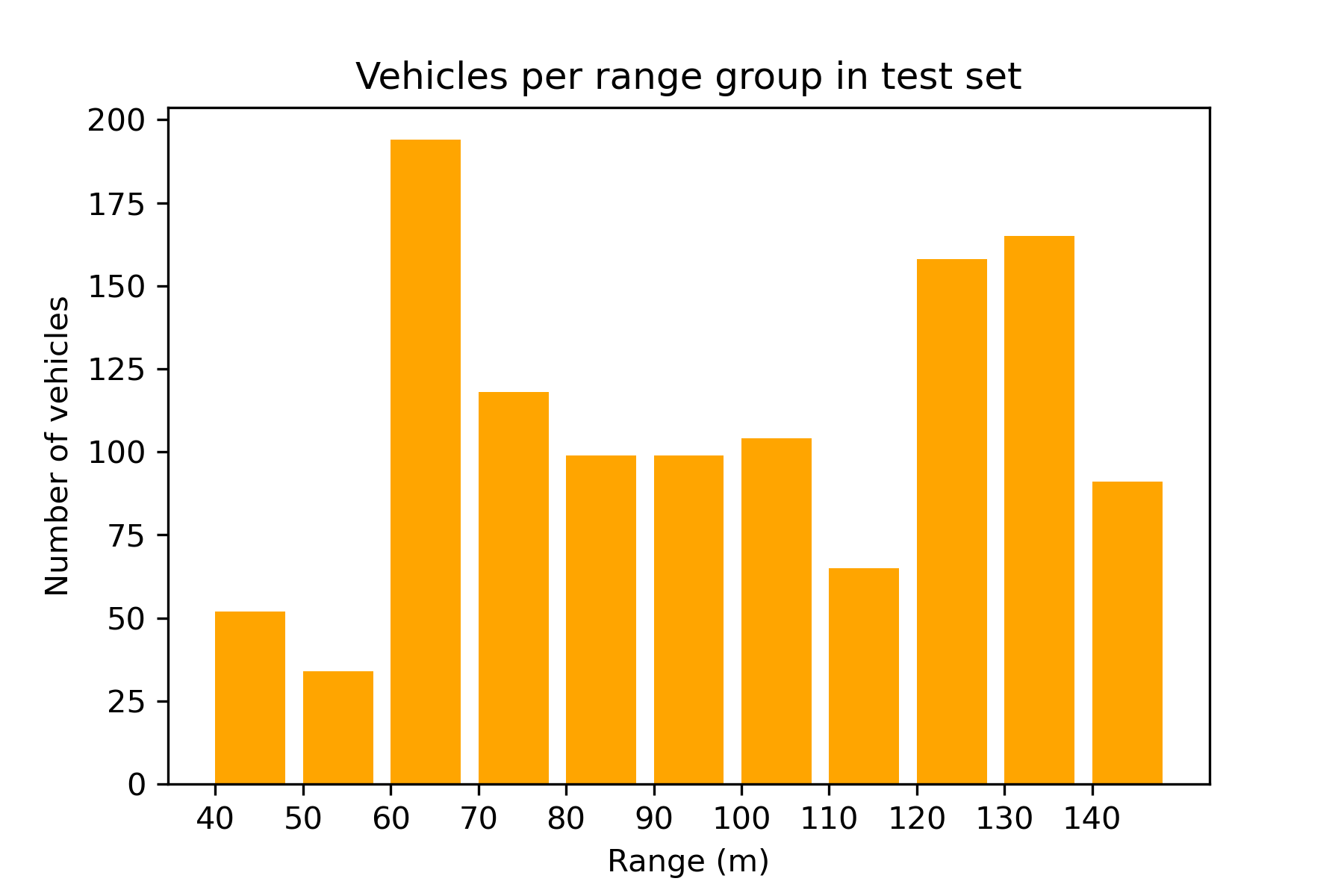}%
    \label{fig_second_case}}
  \caption{Number of vehicle instances by range groups in the train (a) and test (b) sets.}
  \label{fig:train_test}
\end{figure*}

\textit{Output vector}

The output of the 3D object detector is a 22 element tensor with a rich representation of the object's 3D bounding box viewpoint and size. The first 18 elements are the image coordinate pairs of nine keypoints representing the corners and centroid of the 3D box projected on the image plane. It is worthwhile to mention that we introduce a novel normalization step that expresses the keypoint image coordinates in terms of the width of the 2D bounding box (\cref{fig:keypoint}). This forces the model to learn the location of the object's 3D keypoints even when they are out of the 2D image snippet obtained at the first stage. The following three elements are the object's width, length, and height, divided by 10. The final element is the object's observation angle normalized from $[-\pi,\pi]$ to $[0,1]$. The observation angle is that formed between the ray going from the camera center of projection to the object centroid and the object’s X axis, measured in the XZ plane of the camera frame.
Even when all these 22 elements in the output vector are not necessary to determine a 3D bounding box, we keep them because this helps the model to learn useful correlations between the output elements and enforce consistency between their values. Reducing the output vector to fewer elements only decreases the number of parameters by an insignificant amount in the last fully connected layer.

\subsection{Road geometry map}

This component differentiates UrbanNet from previous methods because we do not assume a flat ground, instead, we use a 3D map that contains elevation and slope at any given point on the visible roads to represent their geometry and support the accurate determination of a vehicle pose. More specifically, we suggest a Triangulated Irregular Network (TIN) \cite{958223} as an appropriate data structure to store elevation and slope. The accuracy of the TIN varies depending on the number of input sample points used to build it; therefore, we can easily increase the point density to better represent winding roads or those with drastic changes in slope, where the flat ground approximation does not hold. This approach is also particularly well suited for positioning detected vehicles at long range.

\subsection{Object size and pose}

To obtain the 3D object size and pose we apply the principles of projective geometry and combine the output of the 3D object descriptor with the road geometry map. More specifically, we project three keypoints of the bottom of the 3D bounding box to the road 3D surface and determine the fourth by creating a parallelepiped with the first three, this becomes the base of the 3D bounding box. Next, we use the height of the object as estimated by the descriptor to extrude the base of the 3D box in the direction of the vector normal to the road at that location. The outcome is an accurate 3D bounding box with the corresponding size and pose properly characterized with all three rotation angles.

\section{Experiments and results}
\label{sec:experiments}

\subsection{Synthetic dataset}

We use a synthetic dataset created with specialized rendering software from the Grand Theft Auto V video game. It includes realistic images as taken by long range cameras installed at urban road intersections. The field of view of the virtual long range cameras is 21°, they are positioned at a height of 11m over ground and angled 6° down relative to the horizon. The images have 4k resolution (3840x2160). The dataset labels have the same structure as the KITTI dataset \cite{Geiger2012CVPR}, with three additional attributes that correspond to the object's orientations in X and Z axis as well as the name of the vehicle model. The dataset includes labels for other 3D object classes but we focus on the car and truck classes for our experiments.

\subsection{Model training and evaluation}

To train the 3D object descriptor we use 100 images from each of five camera installations for a total of 500 images. For testing we use 100 images from each of two camera installations for a total of 200 images. The camera installations and viewpoints for the training and test sets are different. \cref{fig:train_test} shows the number of vehicle instances categorized by range in the train and test sets. For data preprocessing we scale and zero pad the image snippets to obtain square tensors of 128 by 128 pixels and scale its values into the $[0,1]$ range. An NVIDIA® GeForce® RTX 2080 Ti is used to train the model for 120 epochs with the Adam optimizer, a batch size of 256, a learning rate of 1e-3, and L1 loss. To evaluate the UrbanNet architecture we use mean average precision (mAP) in bird's eye view (BEV) as this is a standard metric for 3D object detection \cite{Geiger2012CVPR}. In addition, to evaluate the 3D object descriptor independently from the other components of the UrbanNet architecture we measure 3D IoU between each pair of predicted and ground truth 3D bounding boxes.

\subsection{Ablation experiments}

\begin{figure}[!htb]
  \centering
  
  \subfloat[]{\includegraphics[width=.47\textwidth]{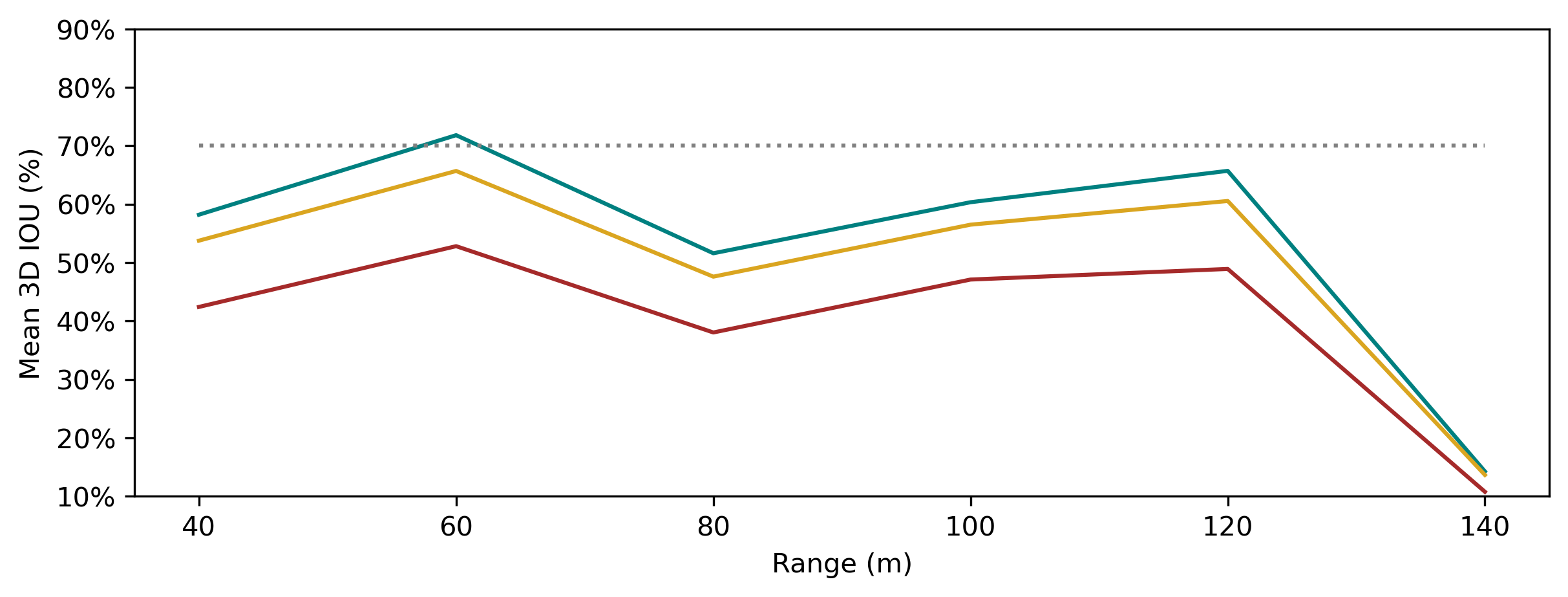}%
    \label{fig_first_case}}
  
  
  \subfloat[]{\includegraphics[width=.47\textwidth]{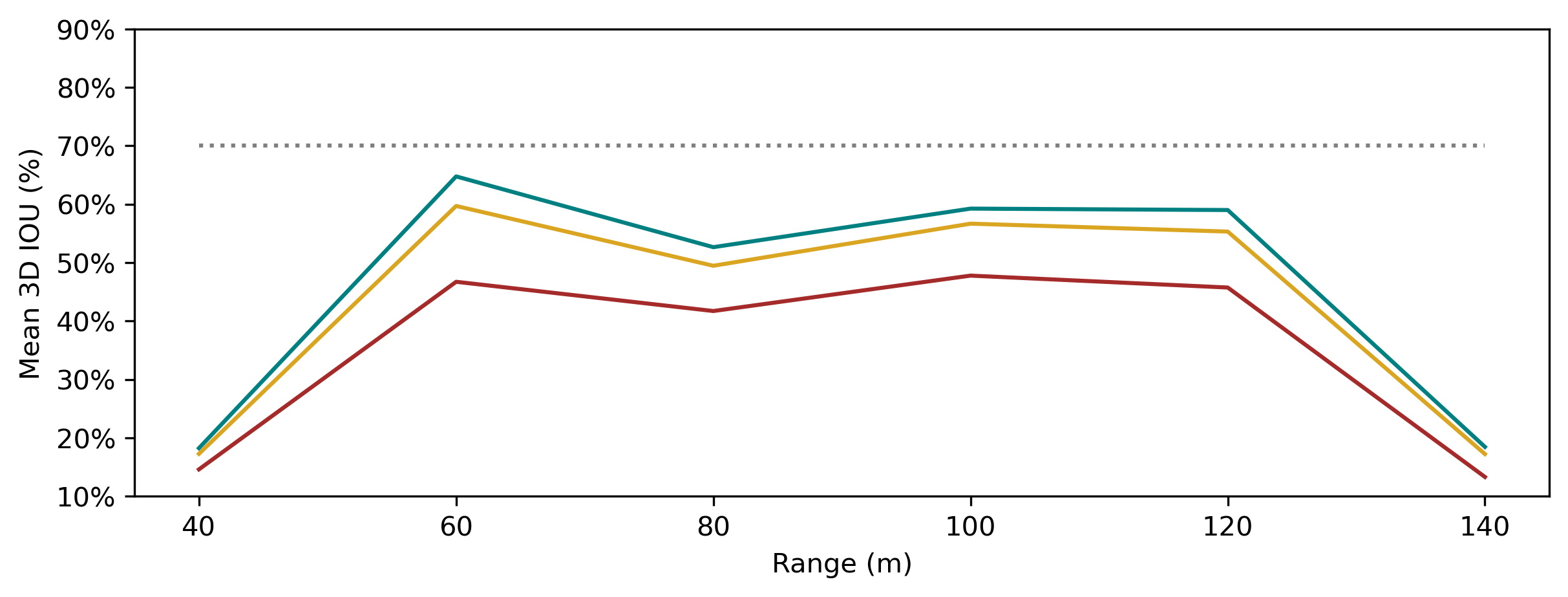}%
    \label{fig_second_case}}
    
  \subfloat[]{\includegraphics[width=.47\textwidth]{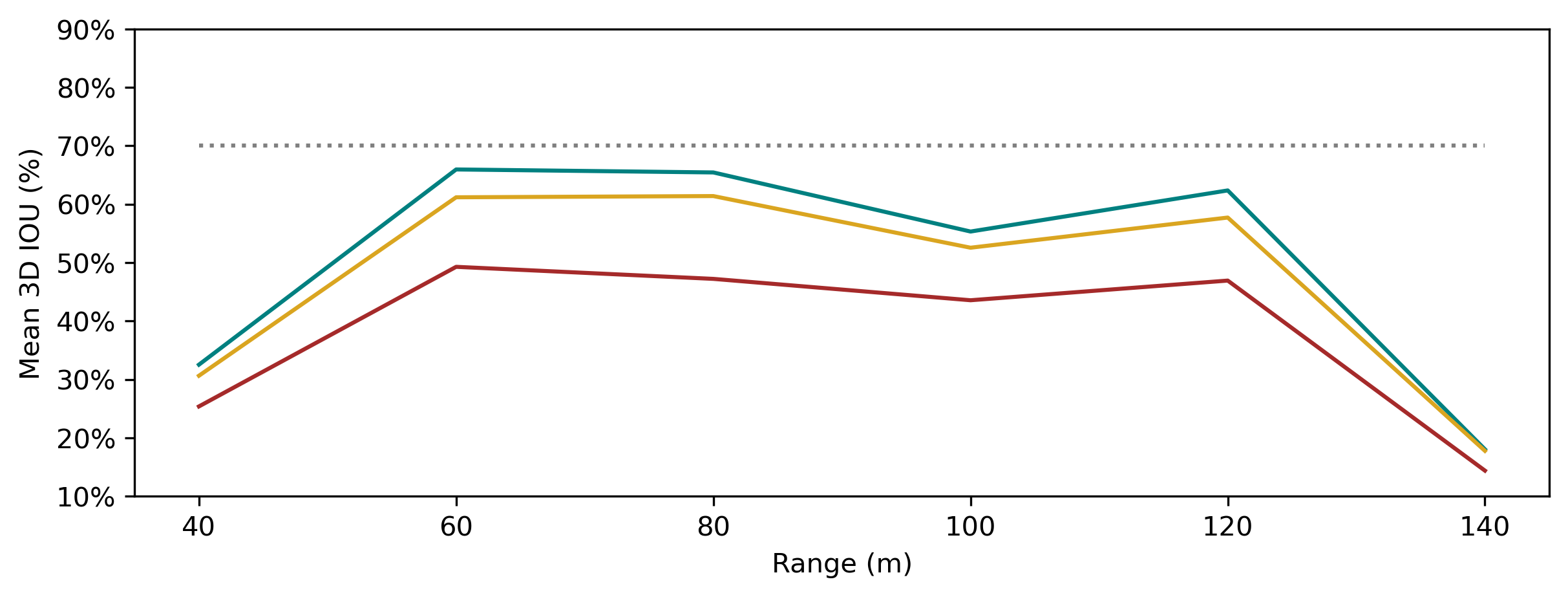}%
    \label{fig_third_case}}
  
  \caption{Mean 3D IoU by range bins in the test set. With (a) and without (b) driving lanes, and predicting only the vertices in the bottom and the height to generate the 3D bounding box (c). These figures also show the effect of adding Gaussian noise in the elevation map where green lines correspond to nominal elevation, orange to a standard deviation of 10cm, and red to a standard deviation of 40cm.}
  \label{fig:oiu_eval}
\end{figure}

The first set of ablation experiments introduce Gaussian noise in the elevation of the road geometry map. The reasoning behind this experiment is that available maps of city roads have varying levels of uncertainty. As-built civil engineering maps documenting infrastructure projects have an accuracy ranging from a few milimeters to several decimeters; therefore, we introduce two levels of Gaussian noise in elevation, one with a standard deviation of 10cm and another with a standard deviation of 40cm (\cref{fig:oiu_eval}). The slope of the map was not perturbed. We see that results are not significantly affected when the noise has a standard deviation of 10cm; on the other hand, they degrade by about 15\% to 20\% levels when the noise reaches 40cm.

The second set of ablation experiments ignores the driving centerlines map as input to the 3D descriptor. For this purpose we set to zero all values in the fourth channel of the image snippet and maintain the same structure of the neural network. The results show that incorporating the lane centerlines certainly contributes to the overall performance of the 3D object descriptor, especially for objects in the range from 40 to 60m. Moreover, even when using the nominal road geometry map the mean IoU values slightly surpass 70\% only for vehicles around 60m from the camera.

In the third set of ablation experiments the 3D object descriptor is trained to predict only the minimal information required to generate 3D bounding boxes. In other words, it only outputs the height and the vertices at the bottom of the box. 

\begin{table}[!htb]
\begin{tabular}{l|r|r|r|}
\cline{2-4}
                                                    & \multicolumn{3}{c|}{\cellcolor[HTML]{EFEFEF}Noise in elevation map}                                                                                                           \\ \hline
\rowcolor[HTML]{EFEFEF} 
\multicolumn{1}{|c|}{\cellcolor[HTML]{EFEFEF}Setup} & \multicolumn{1}{l|}{\cellcolor[HTML]{EFEFEF}Nominal} & \multicolumn{1}{l|}{\cellcolor[HTML]{EFEFEF}STD 10cm} & \multicolumn{1}{l|}{\cellcolor[HTML]{EFEFEF}STD 40cm} \\ \hline
\multicolumn{1}{|l|}{UrbanNet}                       & \textbf{8.069\%}                                       & \textbf{7.533\%}                                        & \textbf{7.031\%}                                        \\ \hline
\multicolumn{1}{|l|}{No driving centerlines}              & 6.622\%                                                & 6.476\%                                                 & 6.075\%                                                 \\ \hline
\multicolumn{1}{|l|}{Keypoints at bottom}           & 7.287\%                                                & 6.974\%                                                 & 6.890\%                                                 \\ \hline
\end{tabular}
\caption{
Average Precision (in \%) measured in bird’s eye view for all the vehicles in the test set. Using a threshold IoU of 0.5 and 40 recall positions.}
\label{tab:map_results}
\end{table}

\cref{tab:map_results} summarizes the bird's eye view results of the ablation experiments in the test set. 
Adding Gaussian noise to the road geometry map affects the performance across all experiments. Predicting only the keypoints at the bottom of the 3D bounding box also reduces the quality of the results but not as much as ignoring the driving centerlines as a prior for the 3D object descriptor. These results are competitive considering the hard challenges we are facing. In UrbanNet we are using a monocular method to estimate boxes oriented in three axes and located at long range.

\subsection{Qualitative results}

\begin{figure}[!htb]
  \centering
  
  \subfloat[]{\includegraphics[width=.25\textwidth]{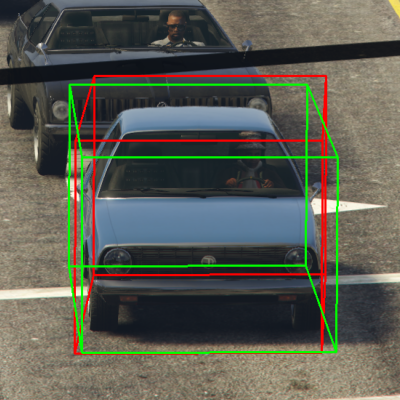}%
    \label{fig_first_case}}
  
  
  \subfloat[]{\includegraphics[width=.25\textwidth]{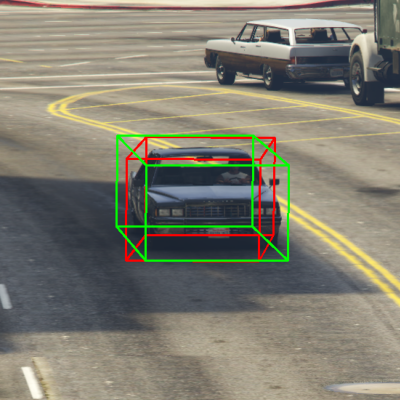}%
    \label{fig_second_case}}
    
  \subfloat[]{\includegraphics[width=.25\textwidth]{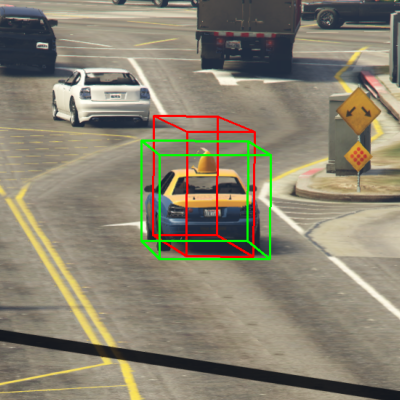}%
    \label{fig_third_case}}
  
  \caption{Examples of detected vehicles at 58m (a), ~109m (b), and ~139m (c) from the camera. Green boxes are ground truth and red correspond to our predictions.}
  \label{fig:qualitative}
\end{figure}

In \cref{fig:qualitative}  we present examples of 3D vehicle detections generated by the UrbanNet architecture. Each example shows the predicted (red) and ground truth (green) 3D bounding box. In general, we see competitive results for vehicles in the range from 60 to 120m.

\subsection{Memory requirements}

The 3D object descriptor has a total of six convolutional layers and three fully connected layers. This neural network architecture comprises 673,902 trainable parameters, a rather small number when compared to other networks with orders of magnitude more parameters. Using the 32-Bit float data type for the network implementation requires less than 3Mb to store all its parameters, this makes the 3D descriptor an efficient model for deployment in edge devices \cite{9138650}\cite{9156225}.



\section{Conclusions}
\label{sec:outlook}

In this paper we introduce UrbanNet, a modular system to perform monocular 3D object detection at long range. Our proposed architecture shows competitive experimental results for accurately detecting vehicles at long range even when they are rotated along their three axes and in roads with changing slope. To the best of our knowledge this is the first attempt to perform full 3D object detection at a range up to 120m. We introduce a novel 3D object descriptor that runs with minimal memory footprint and an innovative use of 3D urban maps to complement the prediction of the vehicles' size and pose. Last but not least important, UrbanNet allows maximum flexibility and interoperability with mature 2D detectors, opening the door to further improvements.

\bibliographystyle{IEEEtran.bst} 
\bibliography{references}

\end{document}